\documentclass[runningheads]{llncs}
\usepackage[margin=3cm]{geometry}   
\usepackage{booktabs}
\usepackage{array}
\usepackage{caption}
\usepackage{float}
\usepackage{cite} 
\usepackage[T1]{fontenc}
\usepackage{graphicx}

\begin{document}
\title{BuildingView: Constructing Urban Building Exteriors Databases with Street View Imagery and Multimodal Large Language Model}
\titlerunning{BuildingView} 
%
%
\author{Zongrong Li\inst{1,2}\orcidID{0009-0008-0933-6439} \and
Yunlei Su\inst{1}\orcidID{0009-0006-8371-7972} \and
Hongrong Wang\inst{1}\orcidID{0009-0005-8867-3940} \and
Wufan Zhao\inst{1}\textsuperscript{*}\orcidID{0000-0002-0265-3465}}
\authorrunning{Li et al.}
%
\institute{
    Thrust of Urban Governance and Design, The Hong Kong University of Science and Technology (Guangzhou), Guangzhou, China\\
    \and
    Spatial Sciences Institute, University of Southern California, Los Angeles, USA \\
\email{zongrong@usc.edu}
\email{ysu186@connect.hkust-gz.edu.cn} 
\email{b1520618977@163.com} 
\email{wufanzhao@hkust-gz.edu.cn} 
}

\maketitle              
\begin{abstract}
Urban Building Exteriors are increasingly important in urban analytics, driven by advancements in Street View Imagery and its integration with urban research. Multimodal Large Language Models (LLMs) offer powerful tools for urban annotation, enabling deeper insights into urban environments. However, challenges remain in creating accurate and detailed urban building exterior databases, identifying critical indicators for energy efficiency, environmental sustainability, and human-centric design, and systematically organizing these indicators. To address these challenges, we propose BuildingView, a novel approach that integrates high-resolution visual data from Google Street View with spatial information from OpenStreetMap via the Overpass API. This research improves the accuracy of urban building exterior data, identifies key sustainability and design indicators, and develops a framework for their extraction and categorization. Our methodology includes a systematic literature review, building and Street View sampling, and annotation using the ChatGPT-4O API. The resulting database, validated with data from New York City, Amsterdam, and Singapore, provides a comprehensive tool for urban studies, supporting informed decision-making in urban planning, architectural design, and environmental policy. The code for BuildingView is available at https://github.com/Jasper0122/BuildingView.

\end{abstract}
\renewcommand{\thefootnote}{\fnsymbol{footnote}}
\footnotetext[1]{Corresponding authors and the first author(Zongrong Li) is a Master's student.}

\section{Introduction}
Urban environments are increasingly becoming the focus of interdisciplinary research, where the integration of spatial and visual data is essential for understanding various aspects of urban life. Within this context, the study of building exteriors has gained significant attention due to its direct impact on critical areas such as energy efficiency\cite{parking-lots}, environmental sustainability\cite{low-carbon}, and human-centric urban design\cite{human-perception}. Specifically, analyzing building exteriors involves extracting detailed spatial and visual features, which are crucial for evaluating urban heat islands, energy consumption, and overall aesthetic coherence. Moreover, building exteriors serve not only as the physical interface between a structure and its environment but also play a crucial role in determining the overall energy performance and aesthetic appeal of urban spaces. 

In this regard, Street View imagery has emerged as an effective tool for extracting detailed features of building exteriors, providing a unique perspective that traditional remote sensing methods often overlook. \cite{liang2024evaluating}. Unlike remote sensing imagery, which provides a top-down view of urban landscapes, street view imagery captures buildings at eye level, revealing intricate details of their spatial attributes, materials, and architectural styles\cite{biljecki2021street}. This granular level of detail is vital for assessing the impact of building exteriors on factors such as urban heat islands, energy consumption, and the overall aesthetic coherence of neighborhoods. Therefore, integrating street view imagery with advanced feature extraction techniques is essential for a comprehensive analysis of building exteriors. While street view imagery offers these advantages, its potential remains underutilized in urban research, particularly when it comes to integrating this visual data with other spatial datasets to provide a comprehensive understanding of building exteriors and their broader urban implications. This highlights the growing need to harness street view imagery more effectively to address the multifaceted challenges of modern urban environments.

Through a systematic search of the urban building exterior research, it is evident that there is a significant gap in the availability of comprehensive databases that integrate street view imagery with detailed building exterior information. Existing studies tend to focus on either the spatial characteristics of buildings—such as their geographic positioning and interactions with surrounding infrastructure—or the aesthetic and material aspects, including innovations in architectural design and material usage\cite{building-style}\cite{material}. However, these perspectives are rarely combined to offer a full understanding of urban building exteriors. The lack of integrated databases that unify spatial data with visual and material information limits the ability to conduct in-depth analyses necessary for informed decision-making in urban planning, architectural design, and environmental policy. Additionally, while recent advancements have leveraged street view imagery for various urban analyses, they primarily focus on environmental factors or general urban features rather than systematically extracting and organizing building exterior information\cite{rundle2011using}. The integration of large language models with street view imagery for detailed urban analysis remains in its nascent stages, highlighting the need for a more integrated approach to understanding urban building exteriors.

To address these knowledge deficits, we propose BuildingView, an innovative approach that integrates Street View imagery with multimodal large language models to create a comprehensive urban building exteriors database. By leveraging high-resolution visual data from Google Street View, combined with spatial and structural data obtained through the Overpass API and OpenStreetMap, this research sets a new standard for the integration of visual and spatial data in urban informatics, significantly advancing the field. Our key contributions include:
\begin{enumerate}
    \item \textbf{Systematic collection and organization of urban building exteriors indicators}, providing a comprehensive understanding of critical factors such as energy efficiency, environmental sustainability, and human-centric design. 
    
    \item \textbf{Development of BuildingView}, a reusable tool that utilizes Street View imagery and multimodal large language models to construct detailed urban building exterior databases for any urban area globally.  
    
    \item \textbf{Construction of extensive urban building exterior databases} for New York, Amsterdam, and Singapore, enabling more informed decision-making in urban planning, architectural design, and environmental policy. 
\end{enumerate}

To outline the structure of this paper, the remainder is organized as follows. \textbf{Section 2} provides a review of relevant literature on building exteriors, discussing both traditional and emerging research in this domain. \textbf{Section 3} introduces our methodological framework, including the systematic literature review, the sampling approach using Overpass and Google Street View APIs, as well as the annotation process with large language models. \textbf{Section 4} presents experimental details and results, followed by a comprehensive analysis of findings. \textbf{Section 5} discusses the implications and limitations of our study. Finally, \textbf{section 6} concludes the paper and offers perspectives for future research.

\section{Related Work}
The current section surveys prior research on building exteriors, discussing the indicators commonly examined in the context of urban informatics, architecture, and sustainability. We divide the related work into two major subsections. \textbf{Subsection 2.1} reviews various building exterior indicators—ranging from structural attributes to energy and environmental aspects—while \textbf{Subsection 2.2} focuses on how Street View Imagery (SVI) and large language models (LLMs) have been leveraged in urban building analysis. This structured overview helps in identifying knowledge gaps that our proposed approach aims to address.

\subsection{Indicators of Building Exteriors Indicators}
In the realm of urban informatics and geoscience, the development of spatiotemporal databases pertaining to buildings has garnered significant attention. Researchers within the disciplines of urban planning and geography typically concentrate on the spatial characteristics of buildings, such as their geographic positioning and the dynamics of their interaction with adjacent infrastructure. For instance, the architectural and urban planning processes must meticulously consider the spatial interplay between buildings and their adjacent parking facilities to optimize energy efficiency \cite{parking-lots}.In contrast, architects and designers are more inclined to examine the intrinsic elements of buildings, including materials and aesthetic design. Recently, the architectural style and the materials used for the exterior of buildings are continually subject to innovation and enhancement \cite{building-style, material}.

The academic discourse has increasingly pivoted towards the examination of the connections between buildings and other indicators, with a particular emphasis on their impact on energy consumption and environmental sustainability. Especially, researchers are actively exploring ways to harness more energy from natural sources through building design, thereby reducing the energy costs associated with buildings. Additionally, they are working to mitigate the thermal impacts caused by the urban heat island effect, which is exacerbated by the heat generated within urban environments \cite{energy}. Concurrently, there is a burgeoning interest in the human-centric aspects of building design, particularly how it fosters engagement and interaction with the public from a participatory perspective \cite{human-perception}.

While numerous open-source databases exist that cater to specific domains of building information, they are often limited to a single aspect of building data. The foundational building attribute knowledge within these databases is typically derived from field surveys or footprint data extracted from remote sensing imagery \cite{chen2022large}. To date, no open-source database has been developed that integrates street view imagery to provide a comprehensive, visually-driven repository encompassing the spatial. Thus, we want to build a spatial database with fundamental, and external surface properties of buildings, with a focus on their relevance to energy, environmental, and human-centric dimensions.

\subsection{SVI and LLMs in Urban Building Analysis}
The advent of street view imagery (SVI) has revolutionized the way urban environments are analyzed, offering a rich and dynamic geospatial data source that rivals traditional remote sensing methods such as satellite imagery \cite{biljecki2021street}. The comprehensive visual data captured by SVI allows for detailed analysis and understanding of urban environments at the street level, providing critical insights into various urban characteristics and phenomena. 

Previous research has demonstrated the utility of SVI in urban studies by integrating it with other urban data sources such as street networks, building information, demographic and socioeconomic data, and survey responses. For instance, SVI has been utilized to audit neighborhood environments \cite{rundle2011using}, and analyze the effects of green view indices on pedestrian activity\cite{ki2021analyzing}. Additionally, studies have explored the use of SVI and machine learning to measure the built environment’s impact on crime \cite{hipp2021measuring}, as well as investigating the relationship between neighborhood walkability factors and walking behaviors using big data approaches \cite{koo2022neighborhood}. Furthermore, SVI has been applied at a global scale, as seen in the creation of extensive datasets that cover millions of street-level images across hundreds of cities, facilitating large-scale urban science and analytics\cite{hou2024global}.

However, to date, there has been no significant work that focuses on using SVI in conjunction with LLMs to construct a comprehensive database of urban building exteriors. The proposed BuildingView project aims to fill this gap by harnessing the combined power of SVI and multimodal LLMs to create an extensive and detailed urban building exteriors database, which can serve as a valuable resource for a wide range of urban studies, including architectural analysis, urban planning, and environmental assessment. 

\section{Method}
\begin{figure}
\includegraphics[width=\textwidth]{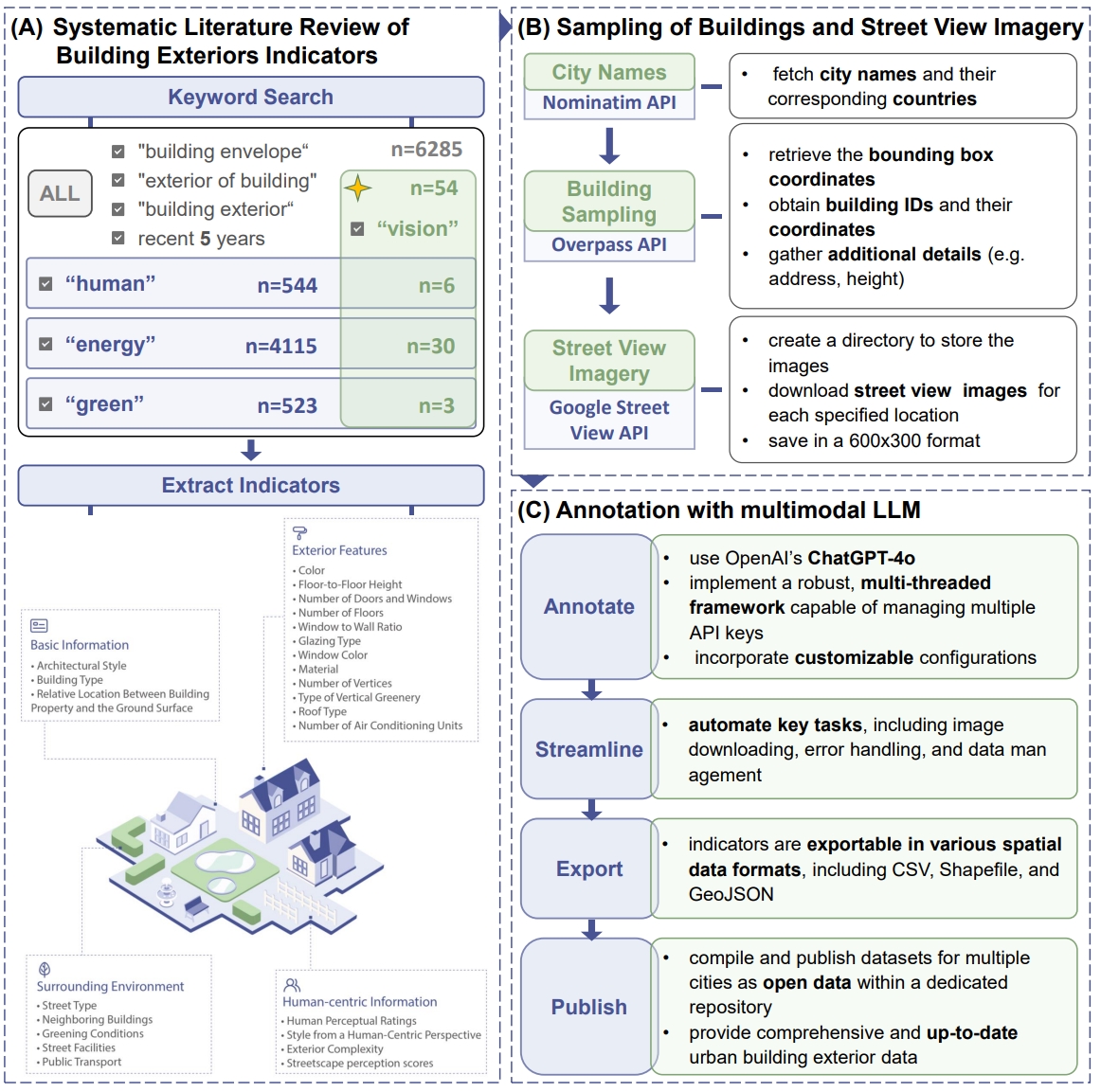}
\caption{BuildingView Workflow: (A) Indicator Review; (B) Sampling; (C) Annotation with ChatGPT-4.0.} \label{fig:workflow}
\end{figure}

In this study, we introduce BuildingView, a novel framework for constructing urban building exterior databases using Street View imagery and multimodal large language models. The research framework is divided into three primary steps, as illustrated in  \verb|Fig| \ref{fig:workflow}, \textbf{subsection 3.1}: a systematic literature review of building exterior indicators; \textbf{subsection 3.2}: sampling of buildings and street view imagery using the Overpass and Google Street View APIs to gather data points; and \textbf{subsection 3.3}: annotation of street view images using the ChatGPT-4O API to create the Urban Building Exteriors Database. Further details on each of these steps are provided below.

\subsection{Systematic Literature Review of Building Exteriors Indicators}
Global building morphology indicators database has concluded many indicators for urban data science \cite{global-indicators}. However, a database of buildings directly covering energy, green, and human-centred buildings has not yet been created. Our goal is to establish a building database that encompasses the majority of indicators of interest to researchers. This database will include comprehensive spatial information on buildings and knowledge extracted from street view images through advanced vision analytics and vision-based reasoning techniques. 

\begin{figure}[h]
    \centering
    \includegraphics[width=0.9\textwidth]{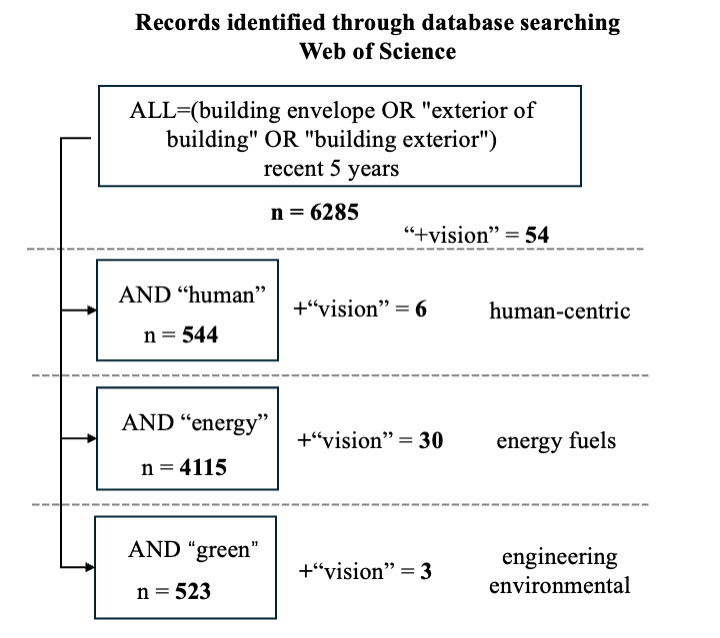}  
    \caption{Literature Review of Building Exteriors}
    \label{fig:literaturereview}
\end{figure}

For that, we have done a systematic literature review on building exteriors \verb|Fig| \ref{fig:literaturereview}. The initial step is to identify emerging topics concerning building exteriors over the past five years. To achieve this, we conduct a keyword search on "building envelope," "exterior of building," and "building exterior" within the Web of Science database. Specifically, this search was performed on November 15, 2024, and restricted to publications from 2019 to 2024, yielding $6,285$ relevant scholarly works. There is a growing discussion about the intelligent design of external surfaces of buildings. The integration of innovative materials and intelligent technologies has the potential to significantly enhance the comfort of indoor environments \cite{wang2024advancements}. 

In addition to fundamental building information and its spatial attributes, we have categorized the concept of "smart building exterior design" into three primary areas that are most frequently discussed by scholars: "human," "energy," and "green." Incorporating these keywords into our search criteria, we have identified a total of $4,683$ relevant works, with the distribution being $544$ for "human," $4,115$ for "energy," and $523$ for "green." 

Given our focus on indicators that can be discerned through visual analysis, we have narrowed down the total literature on building exteriors to $54$ relevant pieces. Within this subset, only $39$ pieces specifically address the hot topic of human-centric, energy fuel and engineering environmental fields. We reviewed these works and we found that among these, the topic of glazing has garnered significant attention. Specific indicators such as glazing type, colour, and Window-Wall Ratio (WWR) are recognized as crucial factors influencing energy absorption \cite{WWR, WWR-and-type}. Outdoor air conditioners, as significant contributors to heat, and their placement methods are also central topics in discussions about energy conservation \cite{aircon-position}. The rooftop, as the uppermost exterior of the building, plays a pivotal role in capturing solar energy. Its design and materials can significantly influence the building's energy efficiency by absorbing and utilizing solar radiation \cite{energy-saving}. From a human-centric perspective, building styles exhibit considerable variation, influenced by factors such as culture, climate, and history \cite{ng2020perception}. They are commonly recognized as falling into categories such as Historical, Somewhat Historical, No Significant Style, Somewhat Modern, and Modern \cite{human-perception}. Vertical greenery not only contributes to aesthetics but also has a positive impact on the thermal insulation of building facades \cite{campiotti2022vertical}. Meanwhile, the surrounding infrastructure, such as transport stations and parking lots, can significantly influence the design of building exteriors and is, in turn, affected by them \cite{transport, parking-lots}. Consequently, we will also consider these elements.

Subsequently, we have identified and summarized the key indicators that researchers find intriguing, which are presented in \verb|Table|. \ref{building_exteriors_indicator_table}. The indicators, which can be derived through vision-based reasoning from street view maps, are predominantly self-explanatory. However, some may not be readily distinguishable by AI technology, and thus, we provide additional explanations for a few that require further clarification. We have considered a range of indicators for our research on building exteriors, encompassing basic building information such as architectural style, building type, and relative location, as well as characteristics like colour. We have also included energy-saving indicators, environmental friendliness indicators—such as Window-Wall Ratio (WWR) and building exterior materials, vertical vegetation type, and human-centric indicators visible on the exteriors, including streetscape perception scores and exterior complexity, etc. Additionally, we have taken into account indicators of the surrounding environment that may influence the design of building exteriors. In parallel, we have integrated useful information for 3D building analysis, including details about neighbouring buildings and greening conditions. Subsequently, we have developed systematic and comprehensive prompts to assist ChatGPT in extracting these indicators from Street View images, aiming to obtain an accurate, detailed, and structured description of the attributes of our building exteriors to fulfill our global spatial database about building exteriors.

\clearpage

\newgeometry{left=1.5cm, right=1.5cm, top=2cm, bottom=2cm}

\begin{table}[p]
    \centering
    \caption{Building Exteriors Indicators}
    \label{building_exteriors_indicator_table}
    \begin{tabular}{p{0.25\textwidth} p{0.4\textwidth} p{0.3\textwidth}}
    \toprule
    \textbf{Indicator} & \textbf{Value Example} & \textbf{Explanation} \\
    \midrule
    Architectural style  & Haussman, Neoclassical, Renaissance, Modernism, Others. &  \\ 
    Building type & single-family houses, multiple-family houses, non-residential buildings & based on the number of units \\ 
    Relative location & on the surface, in the air, underground, across the surface & relative location between building property and the ground surface \\ 
    Colour & red, grey, light brown, unknown. & the colour of the building exteriors \\ 
    Floor-to-Floor height & 3m, approximately 2m, unknown. & inferences based on prior knowledge \\ 
    Number of Doors and Windows & 3 doors and approximately 2 windows, unknown. &  \\ 
    Number of Floors & 2, approximately 2, unknown. & inferences based on prior knowledge \\ 
    WWR & Approximately 0.25, 30\%, unknown. & the ratio of the total area of windows to the total wall area \\ 
    Glazing type & single, double, or triple, unknown & types refer to the number of glass layers in windows \\ 
    Window colour & grey, light brown, unknown. & the colour of the exterior glazing of windows \\ 
    Material & stone, glass product, metallic materials, or indistinguishable due to exterior paint & the material of building exteriors \\ 
    Classification of the building materials & alternative materials, natural materials, or secondary raw materials & the classification based on building exterior materials \\ 
    Number of vertices & - & - \\ 
    Vertical greenery type & panel type, mini planter, cage system box, unknown & the type of vertical greenery system on the building exteriors \\ 
    Roof type & lightweight, green, photovoltaic, vents, rubber & the classification of roof types from an energy-saving perspective \\ 
    Number of air-con units & 3, approximately 3, unknown & number of outdoor air conditioning units \\ 
    Air-con placement type & horizontal, vertical & placement type of outdoor air-con units \\ 
    Street type & residential street, local street, etc. & - \\ 
    Neighbouring buildings & 3 similar buildings, none, etc. & - \\ 
    Greening conditions & 3 trees, 3 grasslands, unknown, etc. & - \\ 
    Street facilities & 3 roads, 3 parking lots, unknown, etc. & - \\ 
    Public transport & 3 subway stations, 3 bus stops, none, unknown, etc. & - \\ 
    Human perceptual ratings & complex, original, ordered, pleasing, boring & - \\ 
    Building style & Historical, somewhat historical, no significant style, somewhat modern, modern & Style of the building from a human-centric perspective \\ 
    Exterior complexity & complex, moderate, simple & inferences based on prior knowledge \\ 
    Streetscape perception scores & Safer, wealthier, livelier, more beautiful, more depressing, more boring & - \\    
    \bottomrule
    \end{tabular}
\end{table}

\restoregeometry
\clearpage

\subsection{Sampling of Buildings and Street View Imagery }
We begin by using the Nominatim API to fetch city names and their corresponding countries based on a query provided through the command line. Nominatim, an open-source geocoding service, outputs a list of cities and countries that match the query, allowing for accurate identification of the desired city and country\cite{clemens2015geocoding}. This makes it easier to directly locate the corresponding bounding box by simply inputting the city and country names, streamlining the data collection process.

For the building sampling, we employ the Overpass API to retrieve and store building data for the specified city and country. The Overpass API \cite{olbricht2011overpass}, which queries the OpenStreetMap (OSM) database, returns building footprints, latitude/longitude coordinates, and occasional attributes (e.g., address, height). While this information is valuable for establishing each building’s spatial context, OSM data generally lacks detailed façade descriptors—such as window-to-wall ratio or materials—that are essential for our analysis. Consequently, we integrate street-view imagery to capture more fine-grained exterior attributes. The Overpass-sourced building records are saved in JSONL format by building types, providing a structured foundation for subsequent merging with street-view data.

Finally, to match the buildings with their corresponding street view exteriors, we use the Google Street View API to download images for the locations specified in the JSONL file. This process involves creating a directory to store the images based on the name of the JSONL file, reading the latitude and longitude coordinates from the file, and downloading images using the Google Street View API for each specified location. To ensure that each downloaded street-view image captures the correct building exterior (rather than facing the opposite side of the street), we adopt the following strategy to determine the camera orientation (heading) when querying the Google Street View API. First, we compute the bearing from the street-view panorama’s GPS coordinate to the building’s centroid. Specifically, we retrieve the panorama’s latitude and longitude returned by the Street View API (the closest panorama to our target building), and then calculate the azimuth angle toward the building’s coordinates. This angle is subsequently used as the heading parameter in our Street View API request, ensuring that the image is oriented to face the building.

Additionally, if no panorama is found within $30$ meters of the building’s coordinate, we expand the search radius in incremental steps (e.g., up to $50$ meters), ensuring a higher likelihood of acquiring at least one valid image in dense urban areas. Through this bearing-based approach, we minimize errors where the camera might otherwise capture the opposite side of the street or irrelevant surroundings. The final images, saved at $600 \times 300$ resolution, are thus more likely to depict the intended building facade.

\subsection{Annotation with Multimodal Large Language Model}
We annotate the street view imagery using OpenAI's ChatGPT-4o, a state-of-the-art language model designed to generate human-like text based on tailored prompts\cite{openai2023chatgpt}. To ensure stability and efficiency in API requests, we implement a robust, multi-threaded framework capable of managing multiple API keys to optimize the distribution of API calls. The prompts for annotation are meticulously developed based on urban building exterior indicators identified during our preliminary research. To maintain flexibility and adaptability in the annotation process, we incorporate customizable configurations, allowing modifications to suit specific research objectives.

We further streamline the annotation process by automating key tasks, including image downloading, error handling, and data management. This automation framework facilitates the execution of tasks according to predefined parameters, manages failed download attempts, consolidates data into a unified dataset while eliminating duplicate records, and tracks processed data to prevent redundancy.

The indicators resulting from our annotation process are designed to be exportable in various spatial data formats, including CSV, Shapefile, and GeoJSON. Our accompanying documentation provides several examples for querying and exporting data into both tabular and geospatial formats, ensuring compatibility with a wide range of analytical tools.

To promote accessibility and facilitate the distribution of ready-to-use datasets, we compile and publish datasets for multiple cities as open data within a dedicated repository. This repository includes datasets for cities and countries with detailed mapping available in OpenStreetMap (OSM), and it continues to expand as we add new locations. The ongoing development of this repository reflects our commitment to providing comprehensive and up-to-date urban building exterior data.

\section{Experiment}
This section describes the experimental setup, including data selection, sampling strategy, and validation of the proposed approach. \textbf{Subsection 4.1} details the data collection from Amsterdam, New York City, and Singapore, along with the associated building and Street View datasets. \textbf{Subsection 4.2} outlines the result analysis and evaluates the performance of our indicators using both manual inspection and quantitative metrics. By examining these diverse case studies, we highlight the strengths and limitations of our method.
\subsection{Study Area and Dataset}
\begin{figure}
\includegraphics[width=\textwidth]{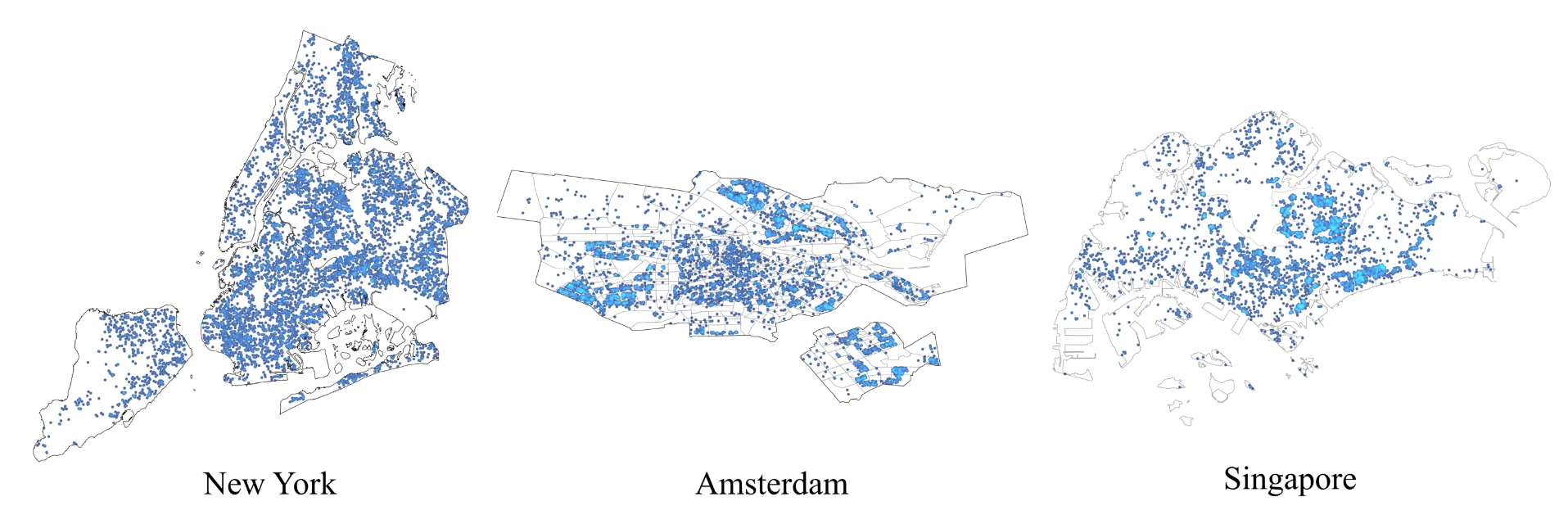}
\caption{The Distribution of Sampling Points} \label{fig:samplingpoints}
\end{figure}

We construct an Urban Building Exteriors Database using Street View Imagery, theoretically applicable to any urban area with street views. For this project, we select three representative cities: New York City (NYC), Amsterdam, and Singapore. The distribution of data points for the three cities is shown in \verb|Fig|\ref{fig:samplingpoints}.

Each city is chosen based on its unique architectural characteristics and urban landscape. New York City is renowned for its iconic skyscrapers and diverse architectural styles, ranging from historic brownstones to modern glass towers\cite{lehan2023city}. Amsterdam offers a distinct contrast with its picturesque canal houses, narrow buildings, and intricate facades, reflecting a historic and compact urban design\cite{havinga2020heritage}. Singapore, known for its modernist high-rise buildings and innovative architectural solutions, represents a dynamic and evolving urban environment that blends tradition with cutting-edge design\cite{yan2022data}.

\begin{figure}
\includegraphics[width=\textwidth]{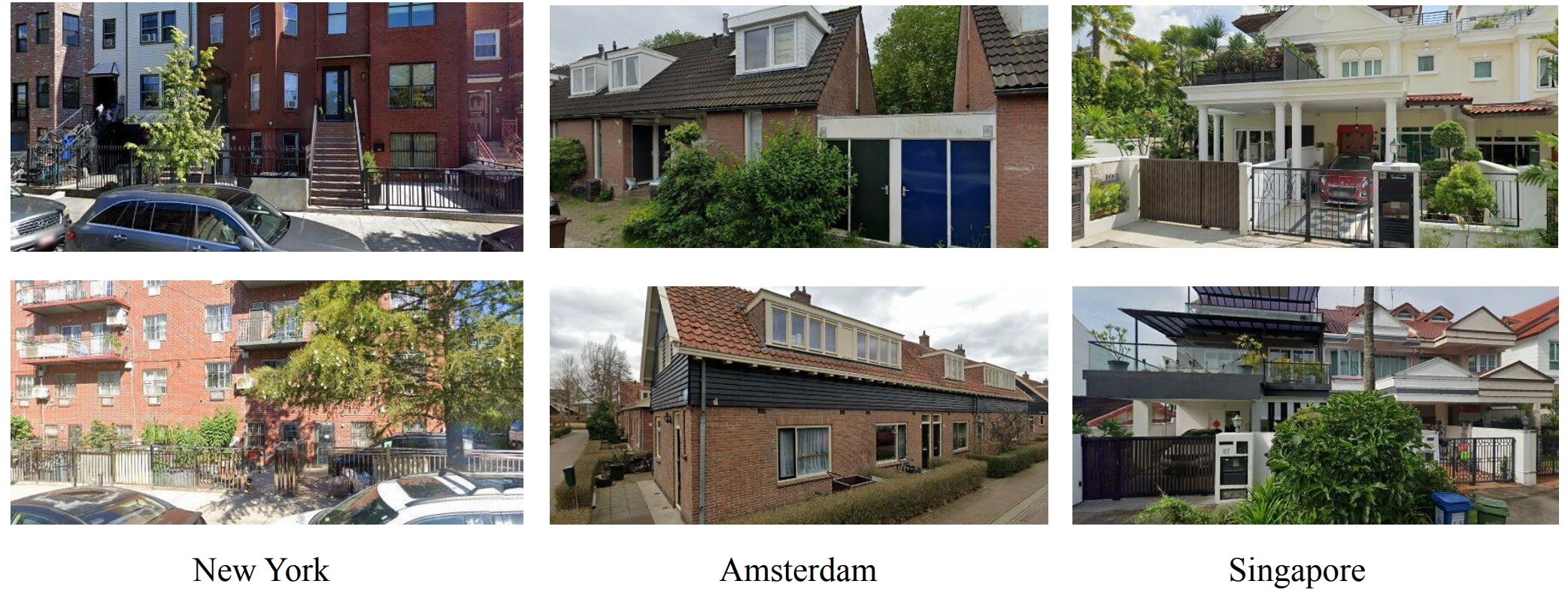}
\caption{Representative Street View Images from Selected Cities} \label{fig:streetviewimages}
\end{figure}

To collect building information, we employ the Overpass API, a powerful tool for querying OpenStreetMap\cite{bennett2010openstreetmap} data, which enables us to compile comprehensive datasets for each city. From these, we sample buildings and collect their building IDs, latitude and longitude coordinates, and address information. For the SVI, we utilize Google Street View Imagery\cite{anguelov2010google} to capture panoramic images of individual building exteriors. To standardize the collection process, we center the building ID coordinates within a 30-meter buffer and select the closest street view image to each building. This method ensures consistency and precision in capturing building exteriors. The dataset comprises $8,130$ building images for NYC, $6,422$ building images for Singapore, and $7,758$ building images for Amsterdam. Additionally, sample street view images for each city have been included to provide visual context for the building exterior data collected, as shown in \verb|Fig| \ref{fig:streetviewimages}.

\subsection{Result and Analysis}
We collected building exterior datasets from Amsterdam, New York City, and Singapore, comprising a total of $22,310$ records. For most extracted indicators, the effective generation rate exceeds 99.90\%. However, the generation rate for streetscape perception scores is relatively lower at 86.05\%. This reduced rate is attributed to the inherent randomness in ChatGPT’s generation process and the challenges associated with complex regular expression matching.

In terms of efficiency, the entire GPT4O-based annotation process across all records incurred a total cost of approximately \$$220$, while covering $26$ indicators per building. This breaks down to roughly \$$0.01$ per building annotation. Each API call required only a few seconds on average. By deploying parallel processing, we managed to complete the entire annotation task within several hours. These results not only highlight the cost-effectiveness of our approach but also demonstrate its potential for scaling to large datasets in urban building data analysis.

To evaluate the model's predictive performance, we select four key variables: Floor-to-Floor Height Numeric Only, Window-to-Wall Ratio (WWR), Vertices, and Tree Coverage. Using manual inspection as the benchmark, we randomly sample 200 instances from the three cities and calculate three key statistical indicators: Mean Absolute Error (MAE), Root Mean Square Error (RMSE), and R². The results show that while predictive accuracy varies among different variables, the overall performance remains strong, as shown in Table \ref{table:performance_metrics}. Notably, the R² values for Floor-to-Floor Height Numeric Only and WWR are $0.83036$ and $0.74679$, respectively. Although the Vertices variable presents certain challenges, its R² value of $0.69767$ still indicates a reasonable level of predictive capability.

\begin{table}[H]  
    \centering
    \caption{Performance Metrics for Different Variables}
    \label{table:performance_metrics}
    \begin{tabular}{p{0.25\textwidth} p{0.15\textwidth} p{0.15\textwidth} p{0.15\textwidth}}
        \toprule
        Variable & MAE & RMSE & R\textsuperscript{2} \\
        \midrule
        Floor-to-Floor Height & 0.01724 & 0.09285 & 0.83036 \\
        WWR & 0.01355 & 0.04130 & 0.74679 \\
        Vertices & 0.82209 & 2.24292 & 0.69767 \\
        Tree & 0.13462 & 0.50000 & 0.66215 \\
        Total & 0.24687 & 0.71927 & 0.73424 \\
        \bottomrule
    \end{tabular}
\end{table}

\begin{figure}
\includegraphics[width=\textwidth]{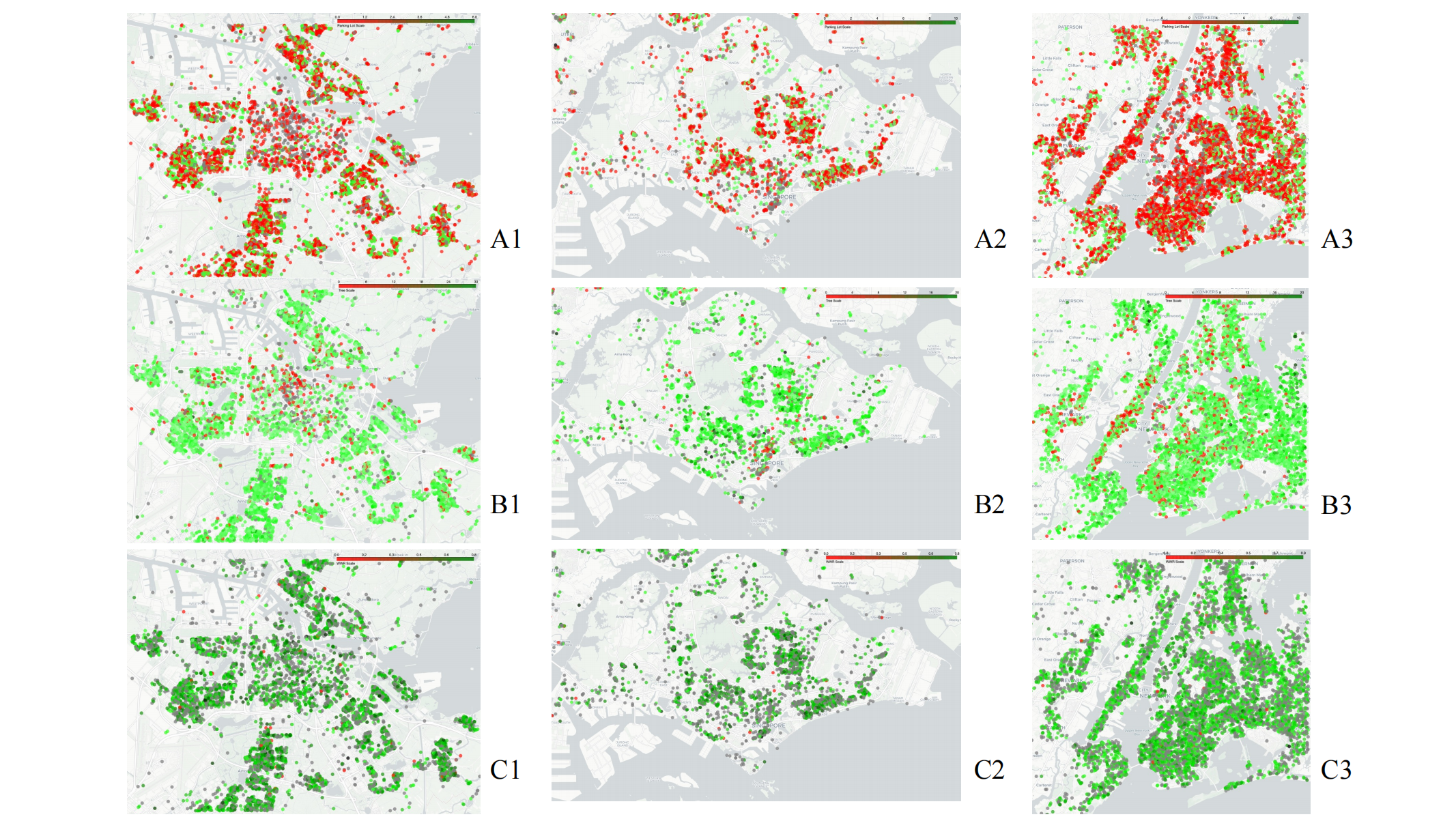}
\caption {Visualization of Result: The Number of Parking Lots (A), Trees (b) and Window-to-Wall Ratio (c)} \label{fig:plot_tree_wwr}
\end{figure}

In the accompanying figures, we select the following indicators from three perspectives—environment, energy, and human—to visualize the results: 'A' represents the parking lot scale, 'B' represents the tree coverage scale, and 'C' represents the Window-to-Wall Ratio (WWR) scale, while the numbers 1, 2, and 3 correspond to Amsterdam, Singapore, and New York City, respectively. In \verb|Fig|\ref{fig:plot_tree_wwr}, the color of the dots transitions from red to green, indicating the scale of the metrics from small to large. Deeper red signifies a smaller scale, while deeper green indicates a larger scale. 

For parking lot scale (A1, A2, A3), deeper red areas in city centers typically indicate smaller or fewer parking lots due to higher population densities and constrained land use, while peripheral or suburban districts appear in lighter or deeper shades of green, suggesting larger or more abundant parking facilities. Tree coverage (B1, B2, B3) shows predominantly green hues, with darker greens indicating denser or more extensive tree presence. Amsterdam maintains balanced tree distribution across its canal networks and suburban edges, Singapore showcases dense greenery in residential neighborhoods (aligned with its garden city initiative), and New York City presents significant variation among boroughs, with some zones rich in street trees and others more built-up. For Window-to-Wall Ratio (C1, C2, C3), intense green signifies high ratios of windows to wall surfaces, commonly found in modern or renovated buildings favoring natural light and large windowpanes, while historic districts display red or lighter green hues due to smaller or fewer windows reflective of older architectural styles.

\begin{figure}[H] 
\centering
\includegraphics[width=0.5\textwidth]{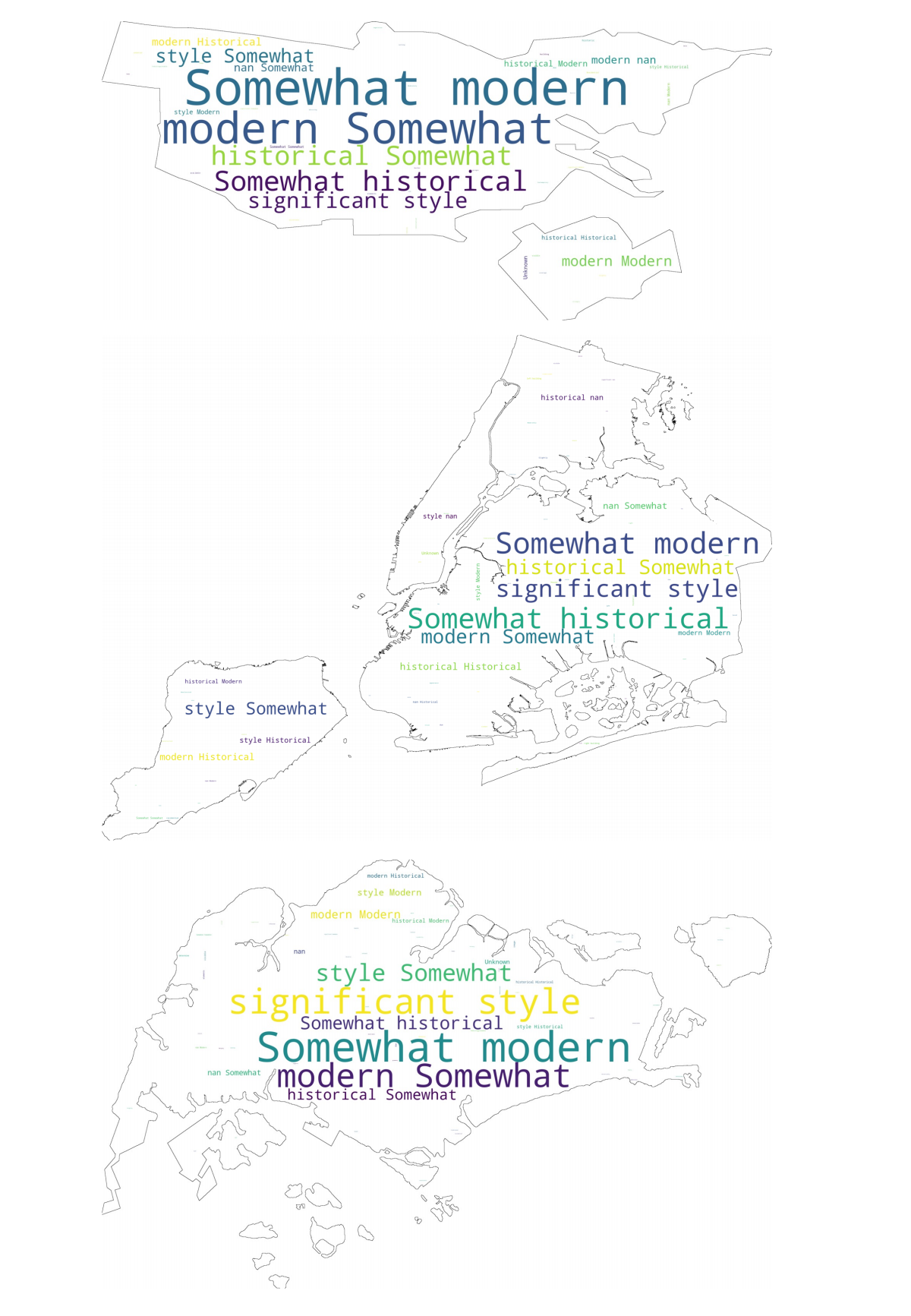}
\caption {The Visualization of Result: Human Style Wordclouds} \label{fig:wordcloud}
\end{figure}

We also create word clouds that capture the diversity of architectural styles across the studied cities. The word clouds reveal that modern and somewhat modern architectural styles are common in some regions, whereas somewhat historical styles prevail in others (see \verb|Fig|\ref{fig:wordcloud}).

The word cloud reveals notable differences in architectural styles across the three cities. In Figures A and C, the terms “somewhat modern” and “modern somewhat” are most prominent, indicating a widespread presence of contemporary architectural designs. This suggests ongoing urban development and renovation activities, which contribute to a modernized cityscape. These areas likely feature more recent constructions with streamlined facades, larger windows, and minimalist aesthetics. In contrast, Figure B shows a strong prevalence of “somewhat historical,” pointing to a significant influence of heritage architecture and preservation efforts. This pattern suggests that historical conservation policies or cultural preferences for traditional styles are more influential in this region. Consequently, the urban landscape here is characterized by older buildings with ornate facades, smaller windows, and classic architectural elements.

Additionally, the term “significant style” appears across all three maps, though its relative prominence varies. In Figures A and C, it suggests distinctive design elements that enhance visual appeal and architectural identity. In Figure B, however, its presence is more subdued, possibly indicating a more uniform historical architectural style. This variation reflects the interplay between modern development and historical preservation across the cities. The word clouds not only highlight stylistic differences but also reveal underlying cultural, historical, and socioeconomic factors that shape urban architectural identity. The diversity in architectural styles underscores how urban planning policies, cultural heritage values, and contemporary design trends contribute to the distinct character of each city.

\section{Discussion}
BuildingView provides an automated, scalable, and open-source framework for extracting building-facade attributes and contextual features from street-view imagery. In what follows, we discuss its value for various fields of built environment research in \textbf{subsection 5.1}, outline the role of sustainability and crowd intelligence in dataset maintenance and enhancement in \textbf{subsection 5.2}, and consider future directions and implementation strategies to further expand its scope and applicability in \textbf{subsection 5.3}.

\subsection{Potential Benefits to Existing Built Environment Research}
\paragraph{Human-Centered Urban Research} Current studies on walkability, aesthetics, and perceived safety are often limited by labor-intensive data collection, reducing research scope and comparability. BuildingView streamlines this by automatically extracting attributes like façade articulation, greenery, window transparency, and pedestrian-scale infrastructure from street-view imagery. These details provide insights into how architectural and environmental elements influence human experiences, such as neighborhood safety and public space quality, supporting evidence-based urban design to enhance walkability and visual appeal.

\paragraph{Scalability and Standardization in Built Environment Data} Built-environment research often lacks standardized, large-scale datasets that capture architectural complexity across different contexts. BuildingView addresses this by systematically extracting and labeling features from street-view images using fine-tuned large language models, ensuring consistent data across regions. Its open-source and modular design allows users to add new indicators or customize the framework, facilitating comparative studies and promoting standardized urban data analytics.

\paragraph{Urban Morphology} Traditional urban morphology studies rely on simplified datasets or field surveys with limited building details. BuildingView enhances this by using street-view imagery to capture detailed façade geometry, window-to-wall ratios, and architectural elements, offering high-resolution morphological indicators. These observations reveal architectural patterns, urban density impacts on livability, and adaptations to cultural or climatic contexts, enabling precise cross-city morphological comparisons and better classification of architectural styles.

\subsection{Sustainability and Crowd Intelligence}
BuildingView is fully automated and reproducible, ensuring continuous updates and responsiveness to urban changes. Scripts for data acquisition, annotation, and enrichment are openly available, enabling scheduled re-runs when new street-view imagery becomes available. As urban environments evolve—new buildings emerge, façades are updated, or land-use patterns change—these automated updates keep the data relevant. Future improvements can integrate advanced building attribute prediction models for greater accuracy and richer metadata.

In addition to automated updates, BuildingView leverages crowd intelligence, turning data users into contributors. In areas with limited street-view coverage, local communities or researchers can enhance the dataset using open mapping platforms like Mapillary or KartaView. An interactive web-based validation tool could also allow volunteers to verify or refine architectural labels, ensuring higher data quality through community participation. This collaborative, open-source approach supports global open-data principles, making BuildingView a valuable resource for urban researchers, planners, and policymakers.

\subsection{Future Directions and Implementation Considerations}
BuildingView is designed to accommodate a variety of urban contexts and building types using street-view data. Ongoing and future enhancements can amplify its robustness, usability, and overall impact. First, broadening geographic reach remains a key priority for capturing buildings in regions where street-view coverage is limited or inconsistent, a process that may benefit from partnerships with local institutions or community-driven imagery initiatives. Second, refining classification algorithms and annotation prompts can help distinguish subtle features and address any class imbalances, especially for rare building elements. Third, volunteer-driven validation tools could allow local users to verify automated labels in challenging scenarios, strengthening data reliability and inclusiveness. Lastly, reducing reliance on proprietary APIs can facilitate more equitable access and sustainability by leveraging open-source or collaborative data platforms where feasible. Through these incremental improvements, BuildingView can serve as a valuable resource for built environment research, policy-making, and community engagement, supporting a clearer understanding of how façades and streetscapes shape the social, environmental, and energy dimensions of cities.

\section{Conclusion}
In this study, we have compiled a list of hot topics related to building exteriors for the last five years. Twenty-six metrics are summarised and categorized into disciplines. Our study presents a novel spatial database for building exteriors, demonstrating the feasibility of integrating AI and street view imagery to enhance the completeness of the database. The validation of our approach highlights its potential for efficiency and accuracy, setting a precedent for similar endeavors. Such a workflow can be replicated for global regions, resulting in a worldwide database of building exteriors. 

Upon enhancing the building classification approach through the integration of multi-source datasets, we anticipate the development of a comprehensive and accurate global building exteriors database. This envisioned database will serve as a solid foundation, providing invaluable insights and data to researchers and industry experts alike.

\begin{credits}
\subsubsection{\ackname} This work is supported by the National Natural Science Foundation of China (Grant No. 42401567), the Tertiary Education Scientific Research Project of the Guangzhou Municipal Education Bureau (Grant No. 2024312159), and the Guangzhou Municipal Science and Technology Bureau Program (Grant No. 2025A03J3640).

\subsubsection{\discintname}
The authors have no competing interests to declare that are 
relevant to the content of this article.
\end{credits}
%
%
%
%
\bibliographystyle{unsrt} 

\end{document}